\documentclass[10pt,twocolumn,letterpaper]{article}

\usepackage{iccv}
\usepackage{times}
\usepackage{epsfig}
\usepackage{graphicx}
\usepackage{amsmath}
\usepackage{amssymb}
\usepackage{xspace}
\usepackage{booktabs}
\usepackage{multirow}
\usepackage[dvipsnames]{xcolor}
\usepackage[leftcaption]{sidecap}
\usepackage{enumitem}
\usepackage{rotating}
\usepackage{dblfloatfix}
\usepackage{listings}
\usepackage{float}

\definecolor{codegreen}{rgb}{0,0.6,0}
\definecolor{codegray}{rgb}{0.5,0.5,0.5}
\definecolor{codepurple}{rgb}{0.58,0,0.82}
\definecolor{backcolour}{rgb}{0.95,0.95,0.92}

\lstdefinestyle{mystyle}{
    backgroundcolor=\color{backcolour},   
    commentstyle=\color{codegreen},
    keywordstyle=\color{magenta},
    numberstyle=\tiny\color{codegray},
    stringstyle=\color{codepurple},
    basicstyle=\ttfamily\scriptsize,
    breakatwhitespace=false,         
    breaklines=true,                 
    captionpos=b,                    
    keepspaces=true,                 
    numbers=left,                    
    numbersep=5pt,                  
    showspaces=false,                
    showstringspaces=false,
    showtabs=false,                  
    tabsize=2,
    showlines=true
}

\newcommand{\sxm}[1]{}

\definecolor{mygrey}{rgb}{0.7, 0.7, 0.7}
\definecolor{mygrey2}{rgb}{0.5, 0.5, 0.5}
\definecolor{mymaroon}{rgb}{0.53, 0.15, 0.34}
\definecolor{mygreen}{rgb}{0.0, 0.6, 0.0}
\definecolor{mygreen}{rgb}{0.0, 0.647, 0.32}
\definecolor{myred1}{hsb}{1, 1., 0.9}
\definecolor{myred2}{hsb}{1, 0.4, 0.9}
\newcommand{\g}[1]{\textcolor{mygrey}{#1}}

\usepackage[T1]{fontenc}

\usepackage{xcolor, soul}
\sethlcolor{white}
\newcommand{\viper}[0]{{\small\fontfamily{txtt}\selectfont \textcolor{mygreen}{\hl{ViperGPT}}}\xspace}
\newcommand{\vipernormal}[0]{{\fontfamily{txtt}\selectfont \textcolor{mygreen}{\hl{ViperGPT}}}\xspace}

\newcommand{\viperemoji}{\includegraphics[height=1\fontcharht\font`\B]{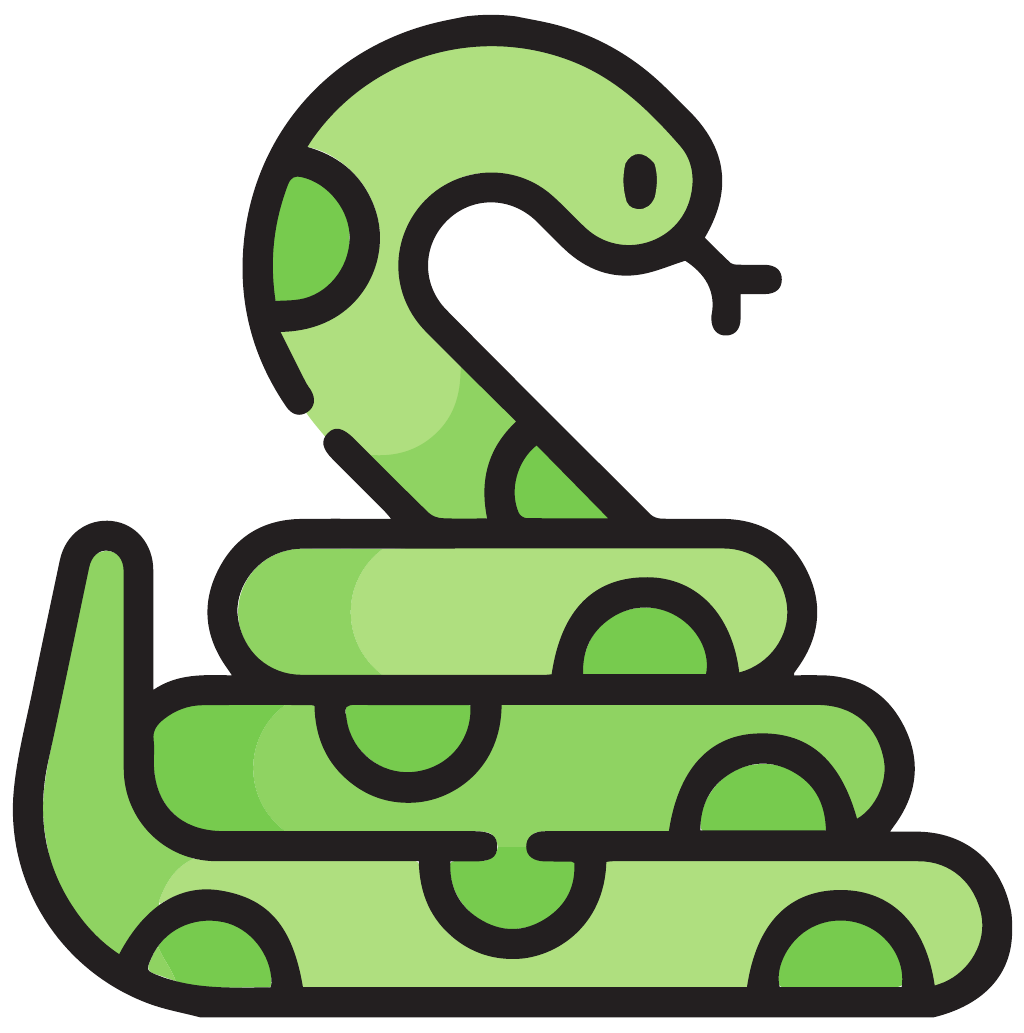}}

\usepackage[scaled=.8]{beramono}  

\usepackage[breaklinks=true,bookmarks=false]{hyperref}

\iccvfinalcopy 

\def\iccvPaperID{11390} 

\ificcvfinal\fi

\everypar{\looseness=-1}

\begin{document}

\title{\vipernormal: Composing Vision and Language via Code}

\title{\vipernormal: Zero-shot Compositions for Multimodal Reasoning}

\title{\vipernormal: a Visual Language Model for Programmatic Reasoning}
\title{\vspace{-0.3cm}\viperemoji\ \vipernormal: Visual Inference via Python Execution for Reasoning with GPT}

\title{\vspace{-0.3cm}\viperemoji\ \vipernormal: Visual Inference via Python Execution for Reasoning}

\title{\vipernormal: Visual Inference via Python Execution for Reasoning}

\makeatletter
\newcommand{\printfnsymbol}[1]{%
  \textsuperscript{\@fnsymbol{#1}}%
}
\makeatother

\author{D\'idac Sur\'is\thanks{Equal contribution. Order determined via coin flip and may be listed either way.}\hspace{0.16cm}, Sachit Menon\printfnsymbol{1}, Carl Vondrick\\Columbia University\\\href{http://viper.cs.columbia.edu}{\texttt{viper.cs.columbia.edu}}
}

\maketitle
\ificcvfinal\thispagestyle{empty}\fi

\begin{abstract}


Answering visual queries is a complex task that requires both visual processing and reasoning.  
End-to-end models, the dominant approach for this task, do not explicitly differentiate between the two, limiting interpretability and generalization. 
Learning modular programs presents a promising alternative, but has proven challenging due to the difficulty of learning both the programs and modules simultaneously.
We introduce \viper, a framework that leverages code-generation models to compose vision-and-language models into subroutines to produce a result for any query. \viper utilizes a provided API to access the available modules, and 
composes them by generating Python code that is later executed.
This simple approach requires no further training, and achieves state-of-the-art results across various complex visual tasks. 
\end{abstract}
\vspace{-0.5cm}

\section{Introduction}

How many muffins can each kid in Figure 1 (top) eat for it to be fair? To answer this, we might 1) find the children and the muffins in the image, 2) count how many there are of each, and 3) reason that `fair' implies an even split, hence divide. People find it natural to compositionally combine individual steps together to understand the visual world. Yet, the dominant approach in the field of computer vision remains end-to-end models, which do not inherently leverage this compositional reasoning. 

Although the field has made large progress on individual tasks such as object recognition and depth estimation, end-to-end approaches to complex tasks must learn to implicitly perform all tasks within the forward pass of a neural network.
Not only does this fail to make use of the advances in fundamental vision tasks at different steps, it does not make use of the fact that computers can perform mathematical operations (\eg, division) easily without machine learning. We cannot trust neural models to generalize systematically to different numbers of muffins or children. End-to-end models also produce fundamentally uninterpretable decisions -- there is no way to audit the result of each step to diagnose failure. As models grow increasingly data and compute-hungry, this approach grows increasingly untenable. We would like to perform new tasks without additional training by recombining our existing models in new ways.

What limits us from creating such modular systems for more complex tasks?
In previous years, the pioneering works of Neural Module Networks \cite{Andreas_2016_CVPR,johnson_inferring_2017,hu_learning_2017}
attempted to decompose tasks into simpler modules. By training end-to-end with modules rearranged in different ways for different problems, the hope was that each module would learn their appropriate function and thereby become reusable. However, numerous issues made this approach difficult to extend to the real world. 
In particular, program generation relied on hand-tuned natural language parsers \cite{Andreas_2016_CVPR}, or otherwise required reinforcement learning from scratch and were thus difficult to optimize \cite{hu_learning_2017,johnson_inferring_2017}. In each case, program generation was highly domain-limited. Furthermore, learning the perceptual models jointly with the program generator made training even more difficult, often failing to produce the intended modular structure \cite{bahdanau_systematic_2019,subramanian_obtaining_2020}.

\begin{figure*}[p]
    \includegraphics[width=\linewidth]{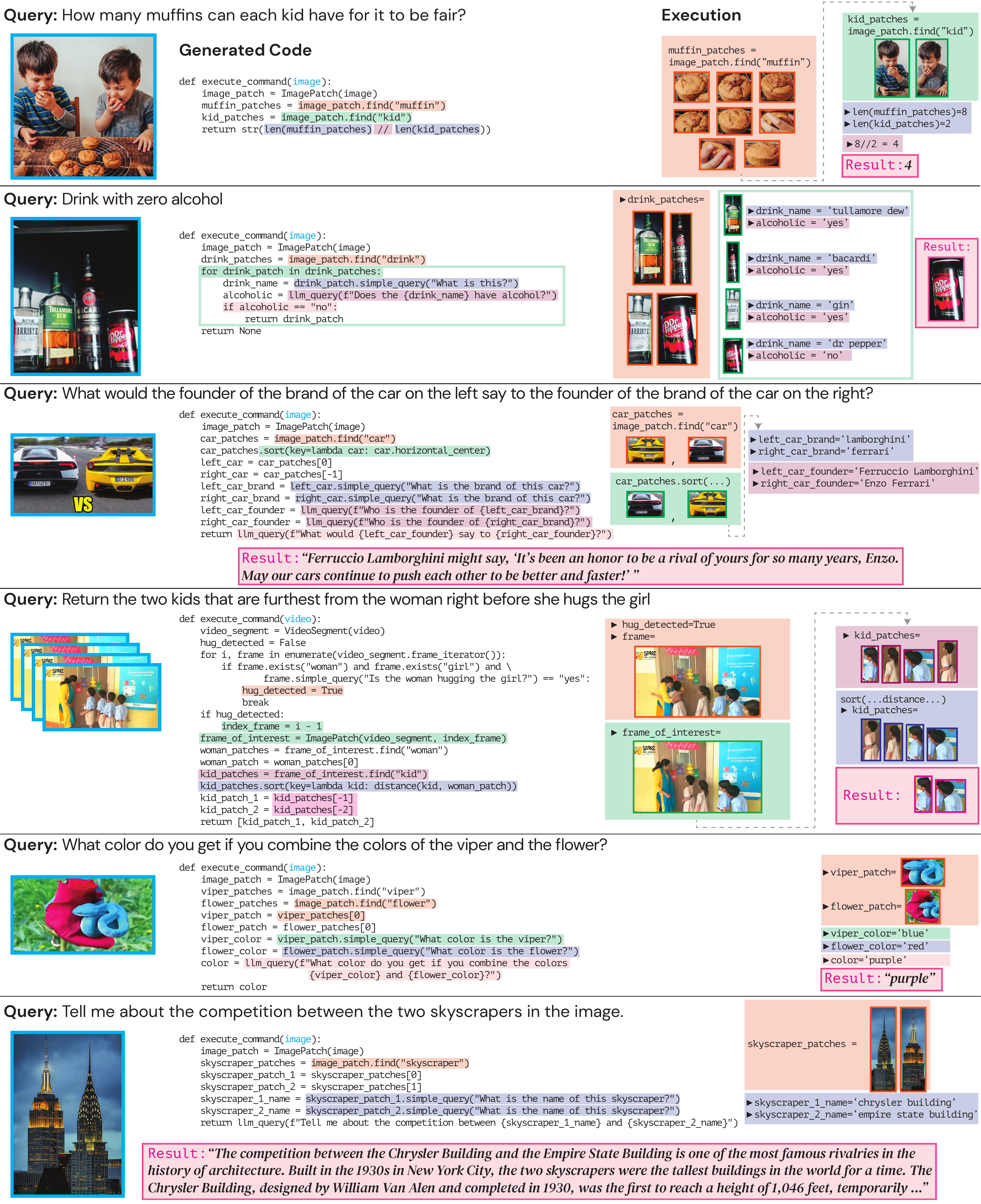}
    \caption{\textbf{In-the-wild results.} Given a visual input and a query, \viper synthesizes a program, then executes it with the Python interpreter in order to produce the final answer. This figure shows both the generated code, and the result of intermediate variables during the execution. By composing pretrained modules, \viper obtains answers that are both correct and interpretable for open-world queries.}
    \label{fig:ood}
\end{figure*}

In this work, we present \viper\footnote{We name our method after a snake because it executes Python code.}, a framework that overcomes these bottlenecks by leveraging  code generating large language models (\eg  GPT-3 Codex \cite{chen2021evaluating}) to flexibly compose vision models based on any textual query that defines the task. It creates customized programs for each query that take images or videos as argument and return the result of the query for that image or video. We show that providing Codex an API exposing various visual capabilities (\eg \texttt{find}, \texttt{compute\_depth}), just as one might provide an engineer, is sufficient for the creation of these programs. The model's prior training on code enables it to reason about how to use these functions and implement the relevant logic. Our results demonstrate that this simple approach delivers remarkable zero-shot performance (\ie without ever training on task specific images).

Our simple approach enjoys many benefits: it is 1) \emph{interpretable}, as all the steps are explicit as code function calls with intermediate values that can be inspected; 2) \emph{logical}, as it explicitly uses built-in Python logical and mathematical operators; 
3) \emph{flexible}, as it can easily incorporate any vision or language module, only requiring the specification of the associated module be added to the API; 4) \emph{compositional}, decomposing tasks into smaller sub-tasks performed step-by-step; 5) \emph{adaptable} to advances in the field, as improvements in any of the used modules will result in a direct improvement in our approach's performance; 6) \emph{training-free}, as it does not require to re-train (or finetune) a new model for every new task; and finally, 7) \emph{general}, as it unifies all tasks into one system.

In summary, our contributions are:
\begin{enumerate}[topsep=0pt,itemsep=-1ex,partopsep=1ex,parsep=1ex]
    \item We propose a simple framework for solving complex visual queries by integrating code-generation models into vision with an API and the Python interpreter, with the benefits above.
    \item We achieve state-of-the-art zero-shot results across tasks in visual grounding, image question answering, and video question-answering, showing this interpretability \textit{aids} performance rather than hindering it.
    \item To promote research in this direction, we develop a Python library enabling rapid development for program synthesis for visual tasks, which will be open-sourced upon publication. 
\end{enumerate}

\section{Related Work}

\textbf{Modular Vision.} Our work takes inspiration from Neural Module Networks \cite{Andreas_2016_CVPR,johnson_inferring_2017}, who argue that complex vision tasks are fundamentally compositional and propose dividing them into atomic perceptual units. This visual reasoning procedure has been explored by a variety of works \cite{kimvisual,Whitehead_2021_CVPR}.
Posterior efforts have focused on explicitly reasoning about the composition by separating the reasoning from the perception, with connections to neuro-symbolic methods \cite{hu_learning_2017,johnson_inferring_2017,yi_neural-symbolic_2018}. These approaches are similar in spirit to ours, but require expensive supervision in the form of programs and end-to-end train the perception modules, which makes them not generalizable to different domains.

Due to the practical difficulty of using these methods, the field has primarily moved towards end-to-end all-in-one models \cite{alayrac2022flamingo,hu2022reveal,huang2023language,li_blip-2_2023}. Such models currently obtain state-of-the-art results, and we compare to them in Section~\ref{sec:experiments}.
Other recent works \cite{zeng2022socraticmodels,reddy2022mumuqa,wang2022language,mao_doubly_2022,menon_visual_2022,gatti2022cofar} show that large pretrained models can be used together to great effect, but hand-specify the particular way models are combined.

Over the course of this project, a surge of interest in the area has resulted in a number of related manuscripts appearing on arXiv which use large language models (LLMs) for automatic module integration. 
In the natural language processing domain, they have been aimed at using external tools \cite{schick2023toolformer,parisi2022talm}, or for structured reasoning using Codex \cite{madaan2022language,wang2022code4struct,gao2022pal,chen2022program}. 
Concurrent work \cite{gupta2022visual} generates a list of pseudocode instructions and interprets them as a `visual program,' relying on in-context learning from provided examples. 
Unlike them, we directly generate unrestricted Python code, which is much more flexible and enables us to demonstrate more advanced emergent abilities, such as control flow and math. Crucially, using Python allows us to leverage the strong prior knowledge Codex learns by training at scale from the Internet. Additionally, we evaluate on many established benchmarks measuring visual understanding and achieve top-performing zero-shot results.


\begin{figure}[t!]
    \centering
    \includegraphics{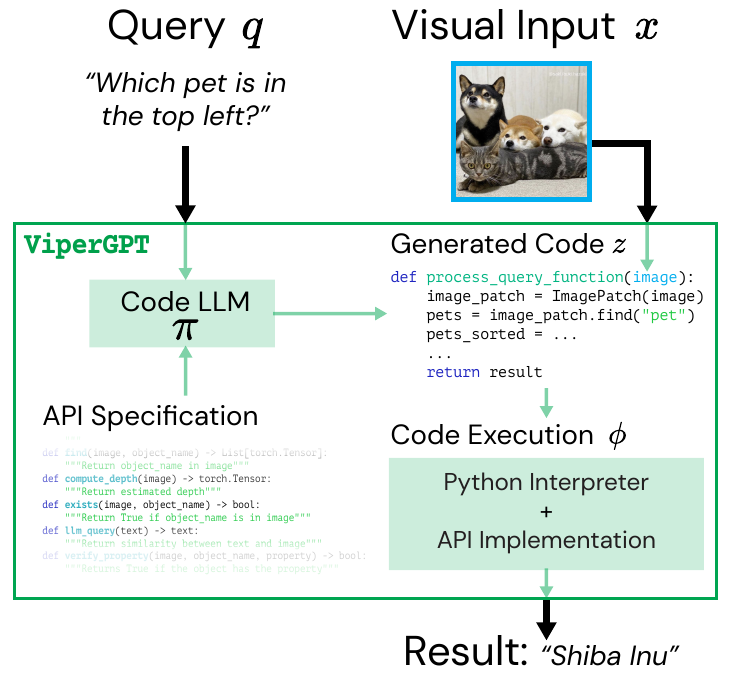}
    \caption{\textbf{Method}. \viper is a framework for solving complex visual queries programmatically.}
    \label{fig:method}
    \vspace{-0.5cm}
\end{figure}


\textbf{Interpretability.} The area of interpretability for complex queries in vision 
is extensive. Many approaches provide explanations in the form of pixel importance, \`a la GradCAM \cite{selvaraju_grad-cam_2020,zhang_interpretable_2018,deng_visual_nodate,park_multimodal_2018}, some also providing textual explanations \cite{park_multimodal_2018}. These are often post-hoc explanations rather than by construction, and do not give step-by-step reasoning including image crops and text. Hard attention in captioning \cite{xu_show_2016} aims for a similar goal regarding intermediate image crops, similarly to our \texttt{find} module, but has proven difficult to incorporate into learning algorithms. See He \etal \cite{he2021interpretable} for a complete overview.

\textbf{Pretrained models.} The perception and external knowledge modules used by \viper are GLIP~\cite{li2022grounded} for object detection, X-VLM~\cite{zeng2021multi} for text-image similarity (as it surpasses CLIP~\cite{radford2021learning} at attribute detection \cite{Bravo_2022_ovad}), MiDaS~\cite{Ranftl2022} for depth estimation, GPT-3~\cite{brown_language_2020} for external knowledge, and \mbox{BLIP-2}~\cite{li_blip-2_2023} for simple visual queries.

\section{Method}\label{sec:method}
We use notation following Johnson \etal\cite{johnson_inferring_2017}.
Given a visual input $x$ and a textual query $q$ about its contents, we first synthesize a program $z = \pi(q)$ with a program generator $\pi$ given the query. We then apply the execution engine $r = \phi(x,z)$ to execute the program $z$ on the input $x$ and produce a result $r$. Our framework is flexible, supporting image or videos as inputs $x$, questions or descriptions as queries $q$, and any type (\eg, text or image crops) as outputs $r$. 

While prior work represents programs as graphs, like syntax trees \cite{johnson_inferring_2017} or dependency graphs \cite{cao_interpretable_2021}, we represent the class of programs $z \in \mathcal{Z}$ directly through Python code, allowing our programs to capitalize on the expressivity and capabilities afforded by modern programming languages.

\subsection{Program Generation}

Johnson \etal\cite{johnson_inferring_2017} and other work in this direction \cite{hu_learning_2017,yi_neural-symbolic_2018,hudson_compositional_2018} typically implement $\pi$ with a neural network that is trained with either supervised or reinforcement learning in order to estimate programs from queries. However, these approaches have largely been unable to scale to in-the-wild settings because either a) the supervision in the form of programs cannot be collected at scale or b) the optimization required for finding the computational graph is prohibitive.

In our approach, we instead capitalize on LLMs for code generation in order to instantiate the program generator $\pi$ that composes vision and language modules together. 
LLMs take as input a tokenized code sequence (``prompt'') and autoregressively predict subsequent tokens. We use Codex \cite{chen2021evaluating}, which has shown remarkable success on code generation tasks. Since we replace the optimization of $\pi$ with an LLM, our approach 
obviates the need for task-specific training for program generation. Using Codex as the program generator and generating code directly in Python allows us to draw on training at scale on the Internet, where Python code is abundant.

\begin{figure}[t]
    \includegraphics[width=\columnwidth]{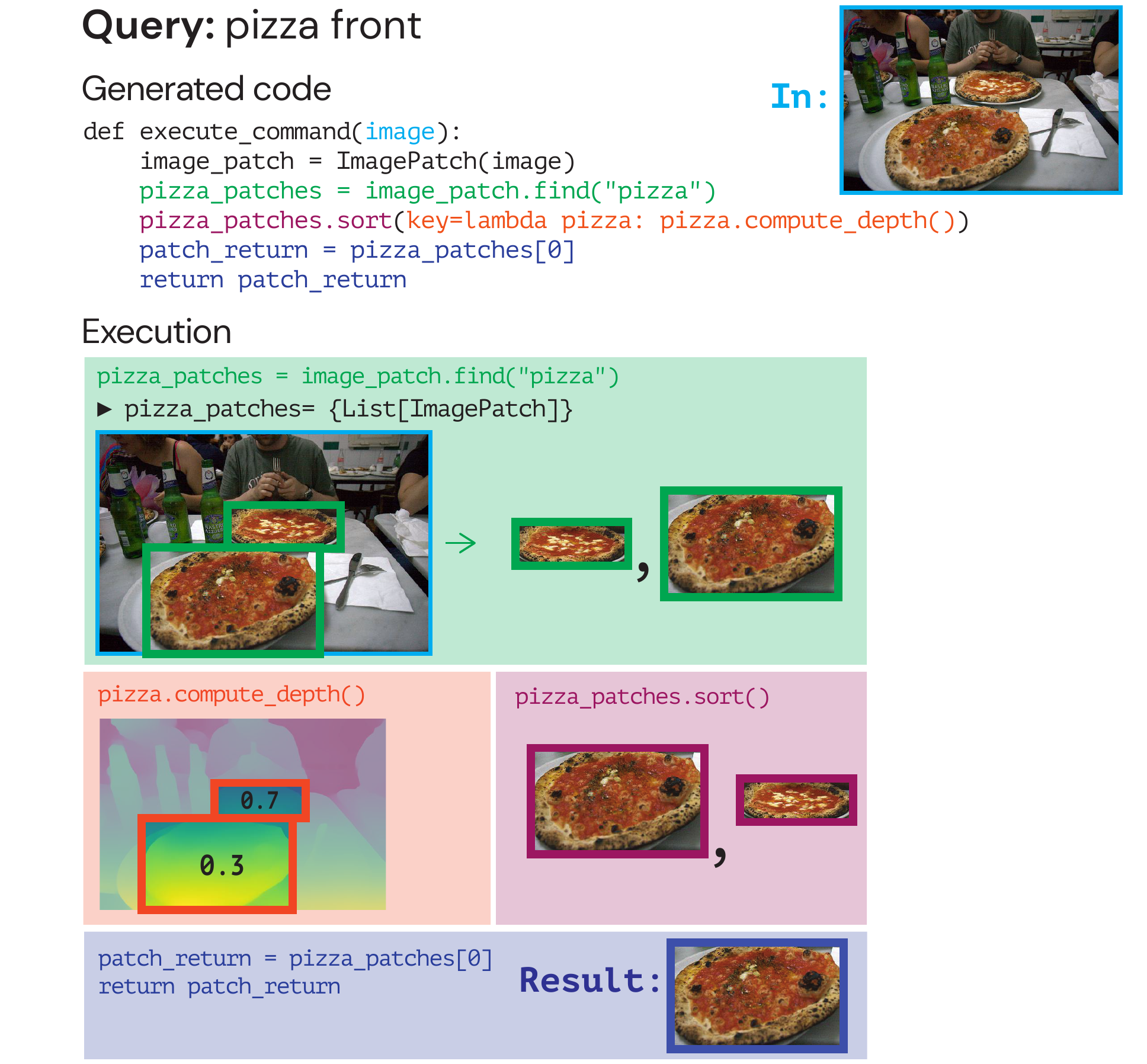}
    \caption{\textbf{Visual grounding on RefCOCO.} 
    }
    \label{fig:refcoco}
\end{figure}



To leverage LLMs in this way, we need to define a prompt that will sample programs $z$ that compose and call these modules as needed. Our prompt consists of an application programming interface (API), detailed in the following section, which we provide to the LLM as part of its input context. 
The final input to the LLM is a sequence of code text consisting of the API specification followed by the query for the sample under consideration. The expected output is a Python function definition as a string, which we then compile and execute.

\begin{table}[t]
    \caption{\textbf{RefCOCO Results}. We report accuracy on the REC task and testA split. ZS=zero shot, \textcolor{mygrey2}{Sup.}=supervised.}
    \label{tab:refcoco_results}
    \centering
        \begin{tabular}{c l c c}
        \toprule
        & &\multicolumn{2}{c}{\textbf{IoU (\%)} $\uparrow$}  \\   
        & & RefCOCO & RefCOCO+\\
        \midrule
        \multirow{2}{*}{\rotatebox[origin=c]{90}{\g{Sup.}}}
        & \g{MDETR \cite{wang2022ofa}}  & \g{90.4} & \g{85.5} \\
        & \g{OFA \cite{wang2022ofa}}  & \g{94.0} & \g{91.7} \\
        \midrule
        \multirow{4}{*}{\rotatebox[origin=c]{90}{ZS}}
        & OWL-ViT \cite{minderer2022simple} & 30.3 & 29.4 \\
        & GLIP \cite{li2022grounded} & 55.0 & 52.2\\
        & ReCLIP \cite{subramanian-etal-2022-reclip} & 58.6 & 60.5 \\
        & \vipernormal (ours) & \textbf{72.0} & \textbf{67.0} \\
        \bottomrule
    \end{tabular}
        \vspace{-0.3cm}
\end{table}

\subsection{Modules and Their API}
\label{sec:api}

Our prompt, included in the Appendix~\ref{sec:appendix_api}, provides the API for different perceptual and knowledge modules, such as for object detection, depth estimation, or language model queries. From this prompt, we found that LLMs are able to induce correct programs $z$ from the query $q$.

\begin{figure*}[t]
    \includegraphics[width=\linewidth]{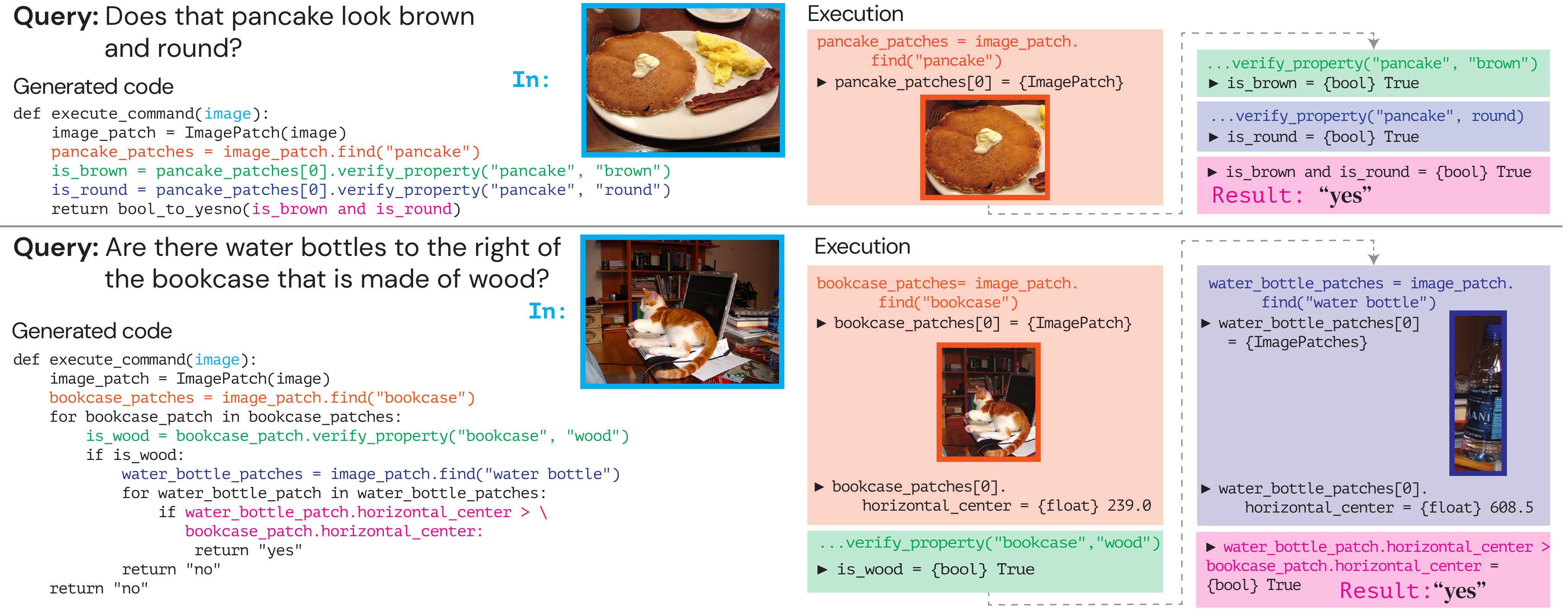}
    \caption{\textbf{Compositional image question answering on GQA.}}
    \label{fig:gqa}
\end{figure*}

The API we provide defines two global classes \texttt{ImagePatch} and \texttt{VideoSegment}, which represent an image patch and a video segment respectively. Each module is implemented as a class method, which internally calls a pretrained model to compute the result. For example, the \texttt{compute\_depth} method of \texttt{ImagePatch} returns an estimate of the median (relative) depth of the pixels in the image patch; we implement this with state-of-the-art large-scale models such as MiDaS~\cite{Ranftl2022}. We provide more details about the modules used in Section~\ref{sec:experiments}. 

The API specifies the input and output types for each method it defines, as well as docstrings to explain the purpose of these functions in natural language. Like most APIs, it additionally provides examples that show how to use these classes and their functions, specified in the form of query-code pairs similarly to in-context learning \cite{suris2020learning,brown_language_2020}. 

The input to Codex does not contain the full \textit{implementation} of the API. Instead, it is given the \textit{specification} for the API, including the function signatures and docstrings. Abstracting away the implementation details is beneficial for two reasons. First, LLM context windows are limited in size \cite{brown_language_2020}, making it infeasible to include the entire implementation. In addition, the abstraction makes code generation independent of changes made to the module implementation.

End-to-end perception modules are excellent when used in the right places, and \viper strongly relies on them. Analogous to dual-system models \cite{kahneman2011thinking} in cognitive science, we argue that generated programs (System 2 - analytic) should be utilized to break down tasks that require multiple steps of reasoning into simpler components, where end-to-end perception modules (System 1 - pattern recognition) are the most effective approach. By composing end-to-end modules into programs, \viper brings the System 2 capability of \textit{sequential processing} to deep learning \cite{bengio_consciousness_2019}.

\subsection{Program Execution}

At execution time, the generated program $z$ accepts an image or video as input and outputs a result $r$ corresponding to the query provided to the LLM. To execute this program, previous work (\eg, \cite{johnson_inferring_2017}) learns an execution engine $\phi$ as a neural module network, composing various modules implemented by neural networks. Their modules are responsible for not only perceptual functions such as \texttt{find}, but also logical ones such as \texttt{compare}. They learn all neural modules together simultaneously end-to-end, which fails to enable systematic generalization \cite{bahdanau_systematic_2019} and results in modules that are not \textit{faithful} to their intended tasks \cite{subramanian_obtaining_2020}, compromising the interpretability of the model.

 We provide a simple, performant alternative by using the Python interpreter in conjunction with modules implemented by large pretrained models. The Python interpreter enables logical operations while the pretrained models enable perceptual ones. Our approach guarantees faithfulness by construction.
 
 The program is run with the Python interpreter; as such, \emph{its execution is a simple Python call}. 
This means it can leverage all built-in Python functions like \texttt{sort}; control flow tools like \texttt{for} or \texttt{if/else}; and modules such as \texttt{datetime} or \texttt{math}. Notably, this does not require a custom interpreter, unlike prior approaches \cite{gupta2022visual,schick2023toolformer} 
Another advantage of a fully Pythonic implementation is compatibility with a wide range of existing tools, such as PyTorch JIT \cite{pytorch_neurips2019}.

\begin{table}
    \caption{\textbf{GQA Results}. We report accuracy on the test-dev set.}
    \label{tab:gqa_results}
    \centering
         \resizebox{0.7\columnwidth}{!}{\begin{tabular}{c l c}
        \toprule
        & &\textbf{Accuracy (\%)} $\uparrow$ \\   
        \midrule
        \multirow{4}{*}{\rotatebox[origin=c]{90}{\g{Sup.}}}
        & \g{LGCN \cite{hu2019language}} & \g{55.8} \\
        & \g{LXMERT \cite{tan2019lxmert}} & \g{60.0} \\
        & \g{NSM \cite{hudson2019learning}}  & \g{63.0} \\
        & \g{CRF \cite{nguyen2022coarse}}  & \g{72.1} \\
        \midrule
        \multirow{2}{*}{\rotatebox[origin=c]{90}{ZS}}
        & BLIP-2 \cite{li_blip-2_2023} & 44.7 \\
        & \vipernormal (ours) & \textbf{48.1} \\
        \bottomrule
    \end{tabular}}
    \vspace{-0.5cm}
\end{table}

In our implementation, each program in a generated batch is run simultaneously with multiprocessing. 
Our producer-consumer design \cite{dijkstra_information_1972} enables efficient GPU batching, reducing the memory and computation costs. Our code is made available at \href{https://viper.cs.columbia.edu/}{\texttt{viper.cs.columbia.edu/}}.

\section{Evaluation}
\label{sec:experiments}
\viper is applicable to any tasks that query visual inputs with text. Unlike other work using large language models for vision tasks, the return values of our programs can be of arbitrary types, such as text, multiple choice selections, or image regions. We select four different evaluation settings to showcase the model's diverse capabilities in varied contexts without additional training. The tasks we consider are: 
1) visual grounding, 
2) compositional image question answering, 
3) external knowledge-dependent image question answering, and 
4) video causal and temporal reasoning.

We consider these tasks to roughly build on one another, with visual grounding being a prerequisite for compositional image question answering and so on. In the following sections, we explore the capabilities \viper demonstrates in order to solve each task.

\subsection{Visual Grounding}


Visual grounding is the task of identifying the bounding box in an image that corresponds best to a given natural language query. 
Visual grounding tasks evaluate reasoning about spatial relationships and visual attributes.
We consider this task first as it serves as the first bridge between text and vision: many tasks require locating complex queries past locating particular objects.

\begin{figure}[t]
    \includegraphics[width=\linewidth]{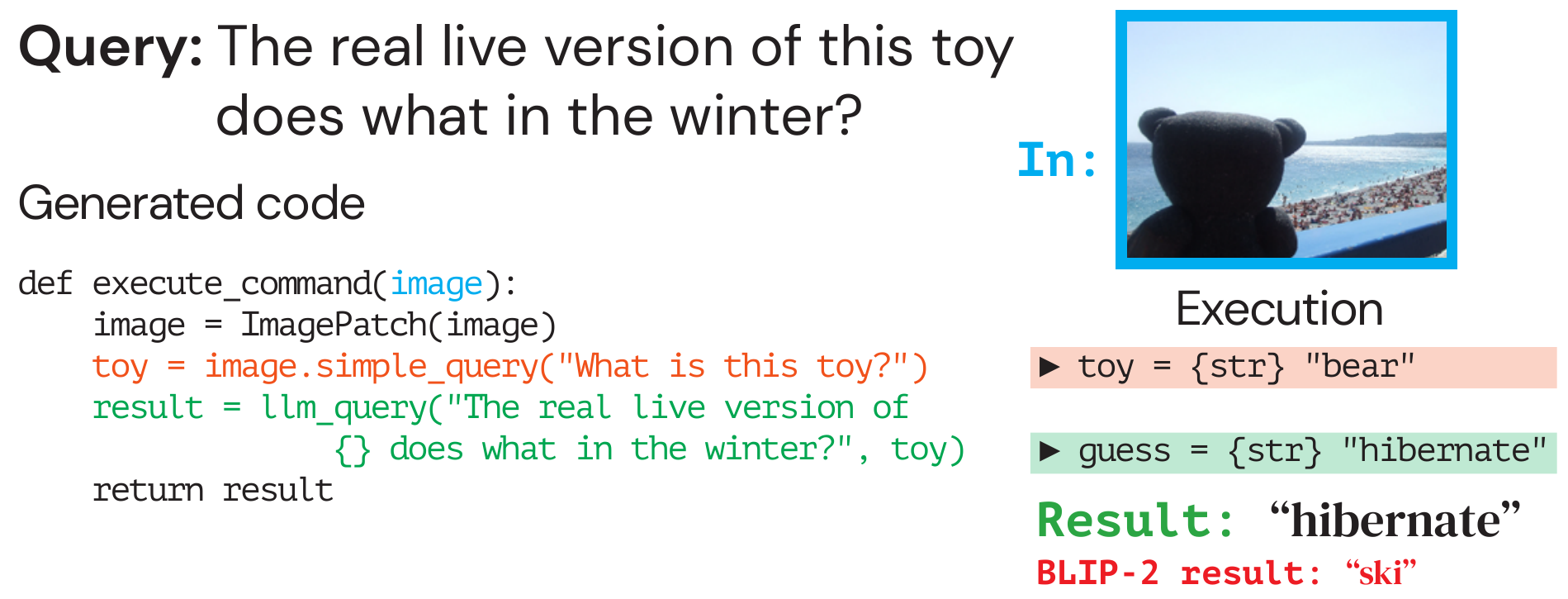}
    \caption{\textbf{Programmatic chain-of-thought with external knowledge for OK-VQA.}}
    \label{fig:okvqa}\vspace{-1em}
\end{figure}

We provide \viper with the  API for the following modules (pretrained models in parentheses). \texttt{\textbf{find}}~({\footnotesize GLIP \cite{li2022grounded}}) takes as input an image and a short noun phrase (\eg ``car'' or ``golden retriever''), and returns a list of image patches containing the noun phrase. \texttt{\textbf{exists}} ({\footnotesize GLIP \cite{li2022grounded}}) takes as input an image and a short noun phrase and returns a boolean indicating whether an instance of that noun phrase is present in the image. Similarly, \texttt{\textbf{verify\_property}} ({\footnotesize X-VLM~\cite{zeng2021multi}}) takes as input an image, a noun phase representing an object, and an attribute representing a property of that object; it returns a boolean indicating whether the property is present in the image. \texttt{\textbf{best\_image\_match}} ({\footnotesize X-VLM~\cite{zeng2021multi}}) takes as input a list of image patches and a short noun phrase, and returns the image patch that best matches the noun phrase. Symmetric to this operation, \texttt{\textbf{best\_text\_match}} takes as input a list of noun phrases and one image, and returns the noun phrase that best matches the image. (This module is not necessary for visual grounding, but rather for tasks with text outputs; we describe it here for simplicity.) They are implemented using an image-text similarity model as in CLIP~\cite{radford2021learning}. Finally, \texttt{\textbf{compute\_depth}} ({\footnotesize MiDaS~\cite{Ranftl2022}}) computes the median depth of the image patch. 
We also define the function \texttt{\textbf{distance}}, which computes the pixel-distance between two patches, using only built-in Python tools.

For evaluation, we use the RefCOCO and RefCOCO+ datasets. The former allows for spatial relations while the latter does not, thereby providing different insights into \viper's capabilities. We compare \viper against end-to-end methods, and outperform other zero-shot methods on both datasets (see Table~\ref{tab:refcoco_results}). We show examples\footnote{Examples in the paper have been cosmetically cleaned by removing comments and error handling, but the logic is unchanged.} in Figure~\ref{fig:refcoco}.
See Appendix~\ref{sec:appendix_models} for more details about the experimental setup.

\begin{figure*}[t]
    \includegraphics[width=\linewidth]{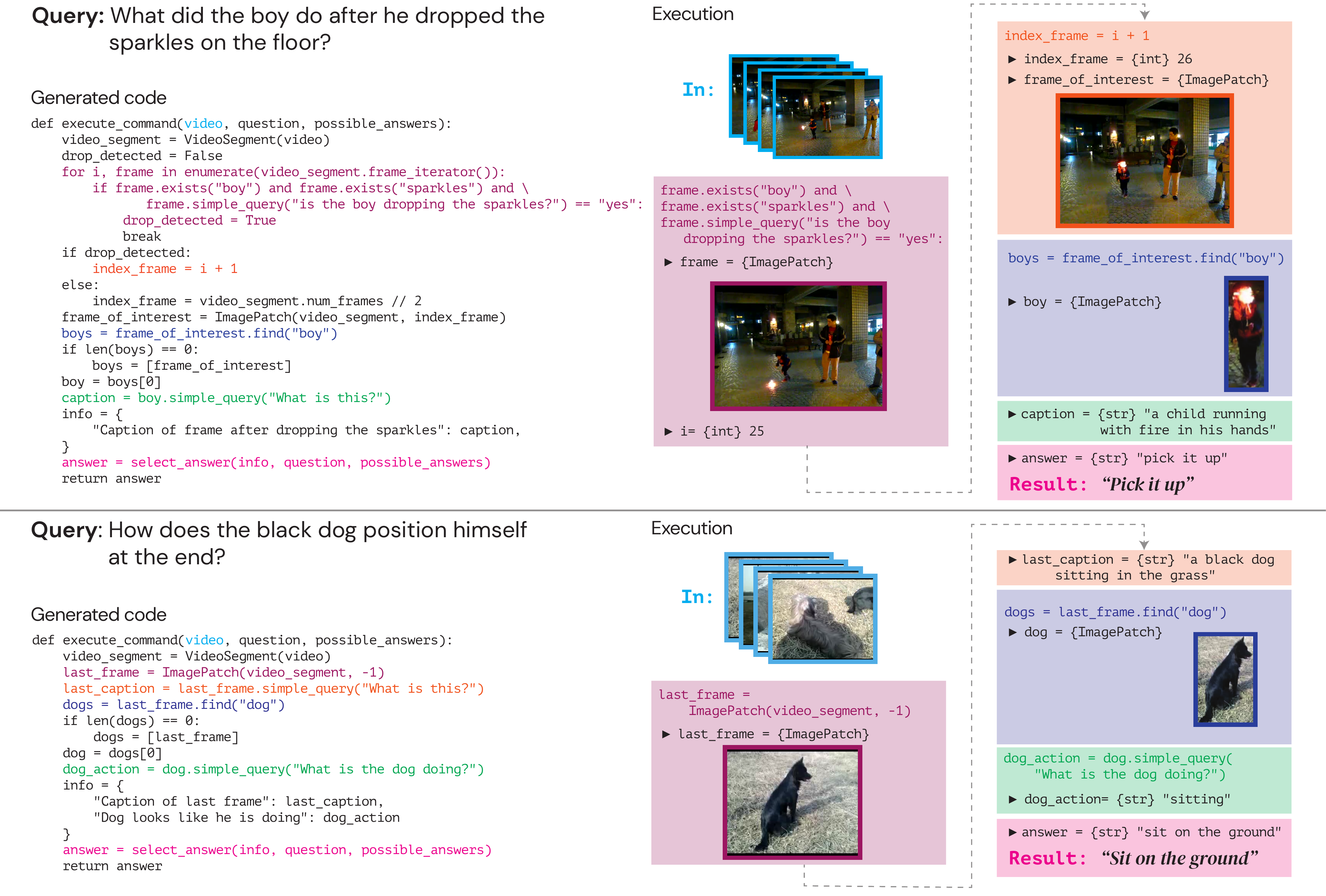}
    \caption{\textbf{Temporal reasoning on NeXT-QA.}}
    \label{fig:nextqa}
\end{figure*}

\subsection{Compositional Image Question Answering}

We also evaluate \viper on image question answering. We focus on compositional question answering, which requires decomposing complex questions into simpler tasks. 
We use the GQA dataset \cite{hudson_gqa_2019}, which was created to measure performance on complex compositional questions. 
Consider Figure \ref{fig:gqa} for example questions as well as our provided reasoning. Even if a question \textit{can} be answered end-to-end, it is both more interpretable and more human-aligned to provide intermediate reasoning rather than requiring the model to compress all steps into one forward pass; as our final result is constructed directly from the intermediate values, they provide a fully faithful interpretation of how the model came to its answer.

For GQA, we incorporate the module \texttt{\textbf{simple\_query}} ({\footnotesize BLIP-2~\cite{li2022grounded}}), which handles basic queries that are not further decomposable, such as ``What animal is this?'' We also add the aforementioned \texttt{\textbf{best\_text\_match}}. This leads us to the best accuracy on GQA among zero-shot models (Table~\ref{fig:gqa}).

\begin{table}
\caption{\textbf{OK-VQA Results}.}
\label{tab:okvqa_results}
\centering
    \begin{tabular}{c l c}
    \toprule
    &&\textbf{Accuracy (\%)} $\uparrow$ \\   
    \midrule
    \multirow{5}{*}{\rotatebox[origin=c]{90}{\g{Sup.}}}
    & \g{TRiG \cite{gao2022transform}} & \g{50.5} \\
    & \g{KAT \cite{gui-etal-2022-kat}} & \g{54.4} \\
    & \g{RA-VQA \cite{lin-byrne-2022-retrieval}} & \g{54.5} \\
    & \g{REVIVE \cite{lin2022revive}}  & \g{58.0} \\
    & \g{PromptCap \cite{hu2022promptcap}}  & \g{58.8} \\
    \midrule
    \multirow{5}{*}{\rotatebox[origin=c]{90}{ZS}}
    & PNP-VQA \cite{tiong-etal-2022-plug} & 35.9 \\
    & PICa \cite{yang2022empirical} & 43.3 \\ 
    & BLIP-2 \cite{li_blip-2_2023} & 45.9 \\
    & Flamingo \cite{alayrac2022flamingo} & 50.6 \\ 
    & \vipernormal (ours) & \textbf{51.9} \\
    \bottomrule
\end{tabular}
\end{table}

\subsection{External Knowledge-dependent Image Question Answering}

Many questions about images can only be answered correctly by integrating outside knowledge about the world. By equipping \viper with a module to query external knowledge bases in natural language, it can combine knowledge with visual reasoning to handle such questions. We add a new module \texttt{\textbf{llm\_query}} ({\footnotesize GPT-3~\cite{brown_language_2020}}), which exploits text models as unstructured knowledge bases. We find that the combination of step-by-step reasoning from Codex along with external knowledge queried from \mbox{GPT-3}'s text model achieves impressive performance in this setting.

We evaluate on the OK-VQA dataset \cite{marino_ok-vqa_2019}, which is designed to evaluate models' ability to answer questions about images that require knowledge that cannot be found in the image. Items in this dataset often require more than one step of reasoning to produce a correct answer. For example, in Figure \ref{fig:okvqa}, one must first perceive from the image that ``this toy'' is a ``bear,'' then use external knowledge to answer what bears do in the winter. End-to-end models must directly produce an answer, and therefore may pick words that are more directly related to the image than the question intended. In this case, the best available end-to-end model guesses ``ski,'' presumably as that is a common winter activity (though, not for bears). \viper, on the other hand, can employ a form of chain-of-thought reasoning \cite{wei_chain_2022} to break down the question as previously described, first determining the type of toy using perception modules and then using the perceived information in conjunction with an external knowledge module to produce the correct response.

\viper outperforms all zero-shot methods, and when compared to models using publicly available resources, it surpasses the best previous model by $6\%$, a wide margin for this dataset (see Table~\ref{tab:okvqa_results}).

\subsection{Video Causal/Temporal Reasoning}

We also evaluate how \viper extends to videos and queries that require causal and temporal reasoning. To explore this, we use the NExT-QA dataset, designed to evaluate video models ability to perform this type of reasoning. 
We evaluate using the NExT-QA multiple choice version.

We provide an additional module \texttt{\textbf{select\_answer}} ({\footnotesize \mbox{GPT-3}~\cite{brown_language_2020}}), which, given textual information about a scene and a list of possible answers, returns the answer that best fits the information. Other than that, the only additional content given in the API is the definition of the class \texttt{VideoSegment}, that contains the video bytestream as well as the start and end timestamps of the video segment that it represents. It also defines an iterator over the frames, which returns an \texttt{ImagePatch} object representing every frame.

\begin{table}
\caption{\textbf{NExT-QA Results}. Our method gets overall state-of-the-art results (including \emph{supervised} models) on the hard split. ``T'' and ``C'' stand for ``temporal'' and ``causal'' questions, respectively.}
\label{tab:nextqa_results}
\centering
    \resizebox{\columnwidth}{!}{\begin{tabular}{c l c c c}
    \toprule
    & &\multicolumn{3}{c}{\textbf{Accuracy (\%)} $\uparrow$}  \\   
    \cmidrule(lr){3-5}
    & & Hard Split - T & Hard Split - C & Full Set\\   
    \midrule
    \multirow{3}{*}{\rotatebox[origin=c]{90}{\g{Sup.}}}
    & \g{ATP \cite{buch2022revisiting}} & \g{45.3} & \g{43.3} & \g{54.3}\\
    & \g{VGT \cite{xiao2022video}}  & \g{-}  & \g{-}& \g{56.9}\\
    & \g{HiTeA \cite{ye2022hitea}}  & \g{48.6} & \g{47.8}& \g{63.1} \\
    \midrule
    \multirow{1}{*}{\rotatebox[origin=c]{90}{ZS}}
    & \vipernormal (ours) & \textbf{49.8} & \textbf{56.4}& 60.0\\
    \bottomrule
\end{tabular}}
\end{table}

We find that despite only being provided with perception modules for images, \viper displays emergent causal and temporal reasoning when applied to videos provided as an ordered list of images. In particular, we observe it generates programs that apply perception to determine which frames are relevant for a given query, then reasons about the information extracted from these frames along with associated frame numbers to produce a final answer. 

Despite seeing no video data whatsoever, \viper achieves accuracy results on par with the best \textit{supervised} model (see Table~\ref{tab:nextqa_results}), and even surpassing it on the NeXT-QA hard split \cite{buch2022revisiting}, both for temporal and causal queries. Of course, the framework of \viper also allows for incorporation of video models, which we expect would further improve the performance well beyond this threshold.

Computational ability presents even more of an obstacle for video understanding than for images. It is infeasible to fit every frame of a moderately-sized video into GPU memory on even the best hardware. \viper may provide a way forward for video understanding that overcomes the limitations of systems that need to perform computation on a whole video simultaneously. See examples in Figure~\ref{fig:nextqa}.

\begin{SCfigure}
  \centering
  \includegraphics[width=0.6\columnwidth]{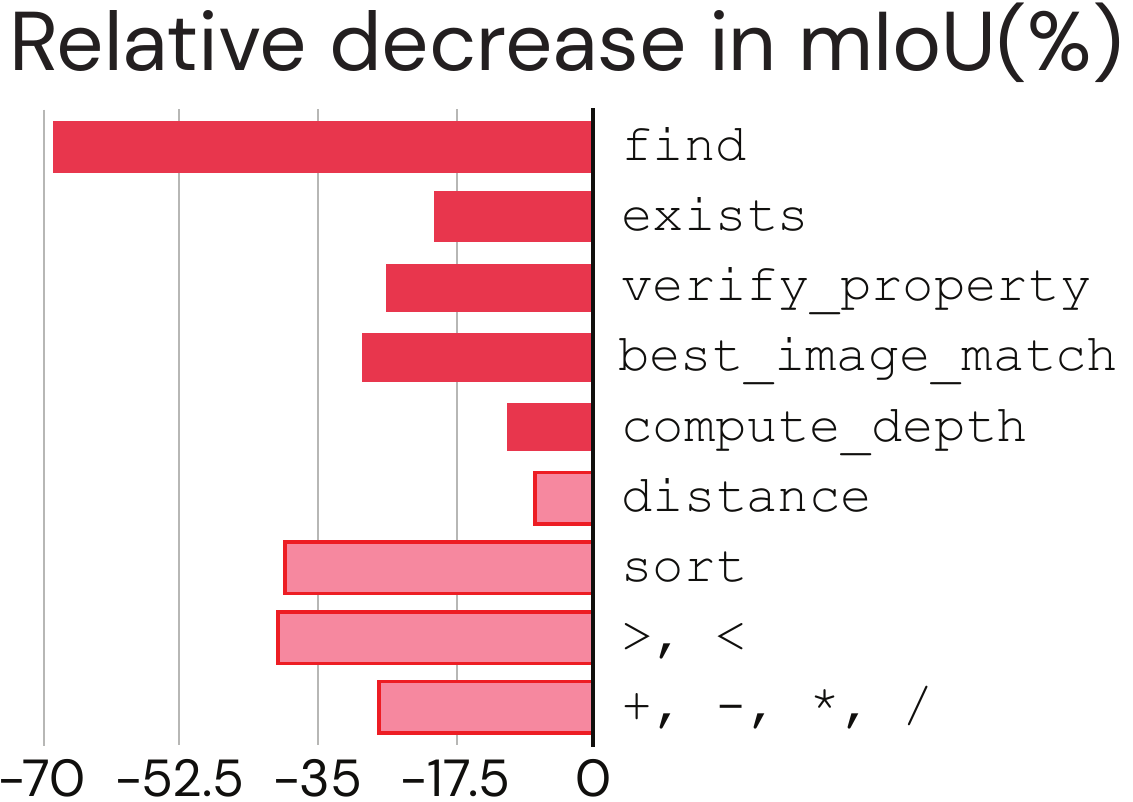}
    \label{fig:intervention}
    \caption{\textbf{Intervention.} We analyze the importance of various \textcolor{myred1}{vision modules} and \textcolor{myred2}{Python functions} in the generated programs as measured by the drop in mIoU when they are made nonfunctional.
  }
\end{SCfigure}

\section{Exploring New Capabilities}

In this section, we showcase various interesting capabilities enabled by use of \viper.

\subsection{Queries Beyond Benchmarks}

We believe that the evident strength of this approach may not be adequately explored by existing benchmarks, which are designed for end-to-end models. In Figure \ref{fig:ood}, we show examples of interesting queries that are interesting in the real world but would not show up in existing benchmarks. We do not add any new API specifications other than the ones already used in the benchmarks. See the Appendix~\ref{sec:appendix_api} for more details.

These examples show that the modules we included are general and cover a wide range of tasks. In settings where new capabilities are required, the framework is general and permits the addition of any modules, like \texttt{ocr}, \texttt{surface\_normal\_estimation}, \texttt{segmentation}, etc.

\subsection{Interventional Explainability}
\looseness=-1
Our programmatic approach enables automatic diagnosis of which modules are responsible for prediction errors, potentially informing which types of models to improve and where to collect more data.  Evaluating the intermediate output of each module is impractical due to the lack of ground truth labels, and naively comparing accuracy
between programs that use a certain module and those that do not could be confounded \eg by the difficulty of the problem.  We can instead perform \textit{interventions} to better understand a module's performance. For each module, we can define a default value that provides no information, and substitute the underlying model for this default output. For instance, \texttt{find} could always return the full input image. We can then consider how much performance drops if evaluating the same code for the examples that use that module. If the intervention has a minimal impact on performance, the module is likely not useful.

\begin{figure}[t]
    \includegraphics[width=\linewidth]{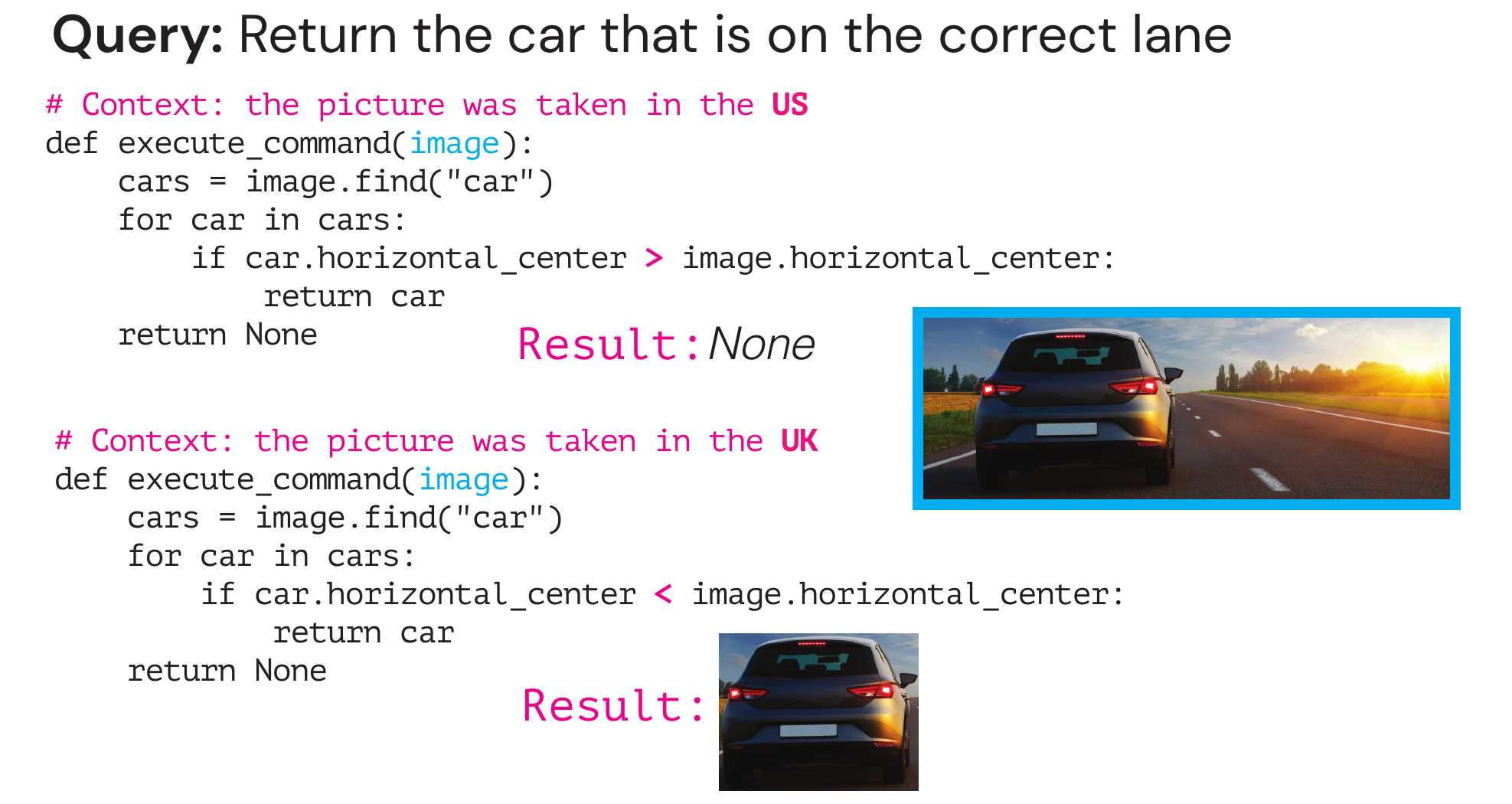}
    \caption{\textbf{Contextual programs}. \viper readily incorporates additional context into the logic of the generated programs.}
    \label{fig:context}
    \vspace{-0.3cm}
\end{figure}

We show an example of this analysis in Figure~\ref{fig:intervention} for visual grounding on RefCOCO, where we observe a similar level of importance for perception modules and Python operations. Both are tightly integrated in our approach.

\subsection{Conditioning on Additional Information}

We found \viper readily admits program generation based on additional knowledge. This context can be provided as a comment prior to the code generation. Such context can be critical to correctly responding to a wide range of queries. In Figure \ref{fig:context} we show one such example. The correct side of the road varies by country, so the initial query cannot be answered. Provided with the context of where the photo was taken, the model produces different logic for each case, adjusted based on the relevant prior knowledge.

\vspace{-0.1cm}
\section{Conclusions}
\vspace{-0.1cm}

We present \viper, a framework for programmatic composition of specialized vision, language, math, and logic functions for complex visual queries. \viper is capable of connecting individual advances in vision and language; it enables them to show capabilities beyond what any individual model can do on its own. As the models implementing these functions continue to improve,  we expect \viper's results will also continue to improve in tandem.

{\small\textbf{Acknowledgements:} This research is based on work partially supported by the DARPA MCS program under Federal Agreement No.\ N660011924032 and the NSF CAREER Award \#2046910. DS is supported by the Microsoft PhD Fellowship and SM is supported by the NSF GRFP.}

{\small
\bibliographystyle{ieee_fullname}
\bibliography{egbib,references}
}

\appendix
\clearpage

\onecolumn 

\section{Pretrained Models}
\label{sec:appendix_models}


We specify details about all the pretrained models used, as well as the code-generation large language model:
\begin{itemize}
    \item \textbf{GLIP \cite{li2022grounded}}. We use the implementation from the official GitHub repository\footnote{https://github.com/microsoft/GLIP}. In our experiments we use the GLIP-L (large) version. In order to adapt to new versions of PyTorch, we had to modify the CUDA implementation of some functions, as the repository relies on old versions of PyTorch. We provide our updated version of GLIP in our code. 
    \item \textbf{MiDaS \cite{Ranftl2022}}. We use the implementation from PyTorch hub\footnote{https://pytorch.org/hub/intelisl\_midas\_v2/}, and use the ``DPT\_Large'' version.
    \item \textbf{BLIP-2~\cite{li_blip-2_2023}}. We tried both the implementation from the official repository\footnote{https://github.com/salesforce/LAVIS/tree/main/projects/blip2} and the Huggingface one\footnote{https://huggingface.co/Salesforce/blip2-flan-t5-xxl}, with little difference between the two, being the former slightly more performant and the latter faster. In both cases, we used the Flan-T5 XXL version.
    \item \textbf{X-VLM~\cite{zeng2021multi}}. We used the official implementation\footnote{https://github.com/zengyan-97/X-VLM}, specifically the version finetuned for retrieval on MSCOCO.
    \item \textbf{GPT-3 for \texttt{llm\_query}}. The GPT-3 model we use for the LLM query function is the \texttt{text-davinci-003} one. We use the official OpenAI Python API\footnote{https://openai.com/blog/openai-api}.
    \item \textbf{Codex}. The GPT-3 model we use for code generation is the \texttt{code-davinci-002} one.
\end{itemize}
See the code for more detailed implementation details.

\section{API}
\label{sec:appendix_api}

We provide the full API next, in Listing~\ref{listing}:

\begin{lstlisting}[language=Python, xleftmargin=.0\textwidth, xrightmargin=.0\textwidth, caption=\textbf{Full API.}, label={listing}]
class ImagePatch:
    """A Python class containing a crop of an image centered around a particular object, as well as relevant information.
    Attributes
    ----------
    cropped_image : array_like
        An array-like of the cropped image taken from the original image.
    left : int
        An int describing the position of the left border of the crop's bounding box in the original image.
    lower : int
        An int describing the position of the bottom border of the crop's bounding box in the original image.
    right : int
        An int describing the position of the right border of the crop's bounding box in the original image.
    upper : int
        An int describing the position of the top border of the crop's bounding box in the original image.

    Methods
    -------
    find(object_name: str)->List[ImagePatch]
        Returns a list of new ImagePatch objects containing crops of the image centered around any objects found in the
        image matching the object_name.
    exists(object_name: str)->bool
        Returns True if the object specified by object_name is found in the image, and False otherwise.
    verify_property(property: str)->bool
        Returns True if the property is met, and False otherwise.
    best_text_match(option_list: List[str], prefix: str)->str
        Returns the string that best matches the image.
    simple_query(question: str=None)->str
        Returns the answer to a basic question asked about the image. If no question is provided, returns the answer
        to "What is this?".
    compute_depth()->float
        Returns the median depth of the image crop.
    crop(left: int, lower: int, right: int, upper: int)->ImagePatch
        Returns a new ImagePatch object containing a crop of the image at the given coordinates.
    """

    def __init__(self, image, left: int=None, lower: int=None, right: int=None, upper: int=None):
        """Initializes an ImagePatch object by cropping the image at the given coordinates and stores the coordinates as attributes.
        If no coordinates are provided, the image is left unmodified, and the coordinates are set to the dimensions of the image.
        Parameters
        -------
        image : array_like
            An array-like of the original image.
        left : int
            An int describing the position of the left border of the crop's bounding box in the original image.
        lower : int
            An int describing the position of the bottom border of the crop's bounding box in the original image.
        right : int
            An int describing the position of the right border of the crop's bounding box in the original image.
        upper : int
            An int describing the position of the top border of the crop's bounding box in the original image.

        """
        if left is None and right is None and upper is None and lower is None:
            self.cropped_image = image
            self.left = 0
            self.lower = 0
            self.right = image.shape[2]  # width
            self.upper = image.shape[1]  # height
        else:
            self.cropped_image = image[:, lower:upper, left:right]
            self.left = left
            self.upper = upper
            self.right = right
            self.lower = lower

        self.width = self.cropped_image.shape[2]
        self.height = self.cropped_image.shape[1]

        self.horizontal_center = (self.left + self.right) / 2
        self.vertical_center = (self.lower + self.upper) / 2

    def find(self, object_name: str) -> List[ImagePatch]:
        """Returns a list of ImagePatch objects matching object_name contained in the crop if any are found.
        Otherwise, returns an empty list.
        Parameters
        ----------
        object_name : str
            the name of the object to be found

        Returns
        -------
        List[ImagePatch]
            a list of ImagePatch objects matching object_name contained in the crop

        Examples
        --------
        >>> # return the children
        >>> def execute_command(image) -> List[ImagePatch]:
        >>>     image_patch = ImagePatch(image)
        >>>     children = image_patch.find("child")
        >>>     return children
        """

    def exists(self, object_name: str) -> bool:
        """Returns True if the object specified by object_name is found in the image, and False otherwise.
        Parameters
        -------
        object_name : str
            A string describing the name of the object to be found in the image.

        Examples
        -------
        >>> # Are there both cakes and gummy bears in the photo?
        >>> def execute_command(image)->str:
        >>>     image_patch = ImagePatch(image)
        >>>     is_cake = image_patch.exists("cake")
        >>>     is_gummy_bear = image_patch.exists("gummy bear")
        >>>     return bool_to_yesno(is_cake and is_gummy_bear)
        """
        return len(self.find(object_name)) > 0

    def verify_property(self, object_name: str, property: str) -> bool:
        """Returns True if the object possesses the property, and False otherwise.
        Differs from 'exists' in that it presupposes the existence of the object specified by object_name, instead checking whether the object possesses the property.
        Parameters
        -------
        object_name : str
            A string describing the name of the object to be found in the image.
        property : str
            A string describing the property to be checked.

        Examples
        -------
        >>> # Do the letters have blue color?
        >>> def execute_command(image) -> str:
        >>>     image_patch = ImagePatch(image)
        >>>     letters_patches = image_patch.find("letters")
        >>>     # Question assumes only one letter patch
        >>>     if len(letters_patches) == 0:
        >>>         # If no letters are found, query the image directly
        >>>         return image_patch.simple_query("Do the letters have blue color?")
        >>>     return bool_to_yesno(letters_patches[0].verify_property("letters", "blue"))
        """
        return verify_property(self.cropped_image, object_name, property)

    def best_text_match(self, option_list: List[str]) -> str:
        """Returns the string that best matches the image.
        Parameters
        -------
        option_list : str
            A list with the names of the different options
        prefix : str
            A string with the prefixes to append to the options

        Examples
        -------
        >>> # Is the cap gold or white?
        >>> def execute_command(image)->str:
        >>>     image_patch = ImagePatch(image)
        >>>     cap_patches = image_patch.find("cap")
        >>>     # Question assumes one cap patch
        >>>     if len(cap_patches) == 0:
        >>>         # If no cap is found, query the image directly
        >>>         return image_patch.simple_query("Is the cap gold or white?")
        >>>     return cap_patches[0].best_text_match(["gold", "white"])
        """
        return best_text_match(self.cropped_image, option_list)

    def simple_query(self, question: str = None) -> str:
        """Returns the answer to a basic question asked about the image. If no question is provided, returns the answer to "What is this?".
        Parameters
        -------
        question : str
            A string describing the question to be asked.

        Examples
        -------

        >>> # Which kind of animal is not eating?
        >>> def execute_command(image) -> str:
        >>>     image_patch = ImagePatch(image)
        >>>     animal_patches = image_patch.find("animal")
        >>>     for animal_patch in animal_patches:
        >>>         if not animal_patch.verify_property("animal", "eating"):
        >>>             return animal_patch.simple_query("What kind of animal is eating?") # crop would include eating so keep it in the query
        >>>     # If no animal is not eating, query the image directly
        >>>     return image_patch.simple_query("Which kind of animal is not eating?")

        >>> # What is in front of the horse?
        >>> # contains a relation (around, next to, on, near, on top of, in front of, behind, etc), so ask directly
        >>> return image_patch.simple_query("What is in front of the horse?")
        >>>
        """
        return simple_qa(self.cropped_image, question)

    def compute_depth(self):
        """Returns the median depth of the image crop
        Parameters
        ----------
        Returns
        -------
        float
            the median depth of the image crop

        Examples
        --------
        >>> # the person furthest away
        >>> def execute_command(image)->ImagePatch:
        >>>     image_patch = ImagePatch(image)
        >>>     person_patches = image_patch.find("person")
        >>>     person_patches.sort(key=lambda person: person.compute_depth())
        >>>     return person_patches[-1]
        """
        depth_map = compute_depth(self.cropped_image)
        return depth_map.median()

    def crop(self, left: int, lower: int, right: int, upper: int) -> ImagePatch:
        """Returns a new ImagePatch cropped from the current ImagePatch.
        Parameters
        -------
        left : int
            The leftmost pixel of the cropped image.
        lower : int
            The lowest pixel of the cropped image.
        right : int
            The rightmost pixel of the cropped image.
        upper : int
            The uppermost pixel of the cropped image.
        -------
        """
        return ImagePatch(self.cropped_image, left, lower, right, upper)

    def overlaps_with(self, left, lower, right, upper):
        """Returns True if a crop with the given coordinates overlaps with this one,
        else False.
        Parameters
        ----------
        left : int
            the left border of the crop to be checked
        lower : int
            the lower border of the crop to be checked
        right : int
            the right border of the crop to be checked
        upper : int
            the upper border of the crop to be checked

        Returns
        -------
        bool
            True if a crop with the given coordinates overlaps with this one, else False

        Examples
        --------
        >>> # black cup on top of the table
        >>> def execute_command(image) -> ImagePatch:
        >>>     image_patch = ImagePatch(image)
        >>>     table_patches = image_patch.find("table")
        >>>     if len(table_patches) == 0:
        >>>         table_patches = [image_patch]  # If no table found, assume the whole image is a table
        >>>     table_patch = table_patches[0]
        >>>     cup_patches = image_patch.find("black cup")
        >>>     for cup in cup_patches:
        >>>         if cup.vertical_center > table_patch.vertical_center
        >>>             return cup
        >>>     return cup_patches[0]  # If no cup found on top of the table, return the first cup found
        """
        return self.left <= right and self.right >= left and self.lower <= upper and self.upper >= lower


def best_image_match(list_patches: List[ImagePatch], content: List[str], return_index=False) -> Union[ImagePatch, int]:
    """Returns the patch most likely to contain the content.
    Parameters
    ----------
    list_patches : List[ImagePatch]
    content : List[str]
        the object of interest
    return_index : bool
        if True, returns the index of the patch most likely to contain the object

    Returns
    -------
    int
        Patch most likely to contain the object

    Examples
    --------
    >>> # Return the man with the hat
    >>> def execute_command(image):
    >>>     image_patch = ImagePatch(image)
    >>>     man_patches = image_patch.find("man")
    >>>     if len(man_patches) == 0:
    >>>         return image_patch
    >>>     hat_man = best_image_match(list_patches=man_patches, content=["hat"])
    >>>     return hat_man

    >>> # Return the woman with the pink scarf and blue pants
    >>> def execute_command(image):
    >>>     image_patch = ImagePatch(image)
    >>>     woman_patches = image_patch.find("woman")
    >>>     if len(woman_patches) == 0:
    >>>         return image_patch
    >>>     woman_most = best_image_match(list_patches=woman_patches, content=["pink scarf", "blue pants"])
    >>>     return woman_most
    """
    return best_image_match(list_patches, content, return_index)


def distance(patch_a: ImagePatch, patch_b: ImagePatch) -> float:
    """
    Returns the distance between the edges of two ImagePatches. If the patches overlap, it returns a negative distance
    corresponding to the negative intersection over union.
    """
    return distance(patch_a, patch_b)


def bool_to_yesno(bool_answer: bool) -> str:
    return "yes" if bool_answer else "no"


def llm_query(question: str) -> str:
    '''Answers a text question using GPT-3. The input question is always a formatted string with a variable in it.

    Parameters
    ----------
    question: str
        the text question to ask. Must not contain any reference to 'the image' or 'the photo', etc.
    '''
    return llm_query(question)


class VideoSegment:
    """A Python class containing a set of frames represented as ImagePatch objects, as well as relevant information.
    Attributes
    ----------
    video : torch.Tensor
        A tensor of the original video.
    start : int
        An int describing the starting frame in this video segment with respect to the original video.
    end : int
        An int describing the ending frame in this video segment with respect to the original video.
    num_frames->int
        An int containing the number of frames in the video segment.

    Methods
    -------
    frame_iterator->Iterator[ImagePatch]
    trim(start, end)->VideoSegment
        Returns a new VideoSegment containing a trimmed version of the original video at the [start, end] segment.
    select_answer(info, question, options)->str
        Returns the answer to the question given the options and additional information.
    """

    def __init__(self, video: torch.Tensor, start: int = None, end: int = None, parent_start=0, queues=None):
        """Initializes a VideoSegment object by trimming the video at the given [start, end] times and stores the
        start and end times as attributes. If no times are provided, the video is left unmodified, and the times are
        set to the beginning and end of the video.

        Parameters
        -------
        video : torch.Tensor
            A tensor of the original video.
        start : int
            An int describing the starting frame in this video segment with respect to the original video.
        end : int
            An int describing the ending frame in this video segment with respect to the original video.
        """

        if start is None and end is None:
            self.trimmed_video = video
            self.start = 0
            self.end = video.shape[0]  # duration
        else:
            self.trimmed_video = video[start:end]
            if start is None:
                start = 0
            if end is None:
                end = video.shape[0]
            self.start = start + parent_start
            self.end = end + parent_start

        self.num_frames = self.trimmed_video.shape[0]

    def frame_iterator(self) -> Iterator[ImagePatch]:
        """Returns an iterator over the frames in the video segment."""
        for i in range(self.num_frames):
            yield ImagePatch(self.trimmed_video[i], self.start + i)

    def trim(self, start: Union[int, None] = None, end: Union[int, None] = None) -> VideoSegment:
        """Returns a new VideoSegment containing a trimmed version of the original video at the [start, end]
        segment.

        Parameters
        ----------
        start : Union[int, None]
            An int describing the starting frame in this video segment with respect to the original video.
        end : Union[int, None]
            An int describing the ending frame in this video segment with respect to the original video.

        Examples
        --------
        >>> # Return the second half of the video
        >>> def execute_command(video):
        >>>     video_segment = VideoSegment(video)
        >>>     video_second_half = video_segment.trim(video_segment.num_frames // 2, video_segment.num_frames)
        >>>     return video_second_half
        """
        if start is not None:
            start = max(start, 0)
        if end is not None:
            end = min(end, self.num_frames)

        return VideoSegment(self.trimmed_video, start, end, self.start)

    def select_answer(self, info: dict, question: str, options: List[str]) -> str:
        return select_answer(self.trimmed_video, info, question, options)

    def __repr__(self):
        return "VideoSegment({}, {})".format(self.start, self.end)
\end{lstlisting}

Not all methods are used in all the benchmarks. Next we describe in more detail what content is used for the API specifications for every benchmark.
\begin{itemize}
    \item \textbf{RefCOCO and RefCOCO+}. We use all the methods from the \texttt{ImagePatch} class except for \texttt{best\_text\_match} and \texttt{simple\_query}. We also use the \texttt{best\_text\_match} and \texttt{distance} functions. Additionally we add \texttt{ImagePatch} usage examples in the API definition that are representative of the RefCOCO dataset, and look like the following:
    \begin{lstlisting}[language=Python, caption=\textbf{RefCOCO example.}, xleftmargin=.0\textwidth, xrightmargin=.0\textwidth]
# chair at the front
def execute_command(image) -> ImagePatch:
    # Return the chair
    image_patch = ImagePatch(image)
    chair_patches = image_patch.find("chair")
    chair_patches.sort(key=lambda chair: chair.compute_depth())
    chair_patch = chair_patches[0]
    # Remember: return the chair
    return chair_patch
\end{lstlisting}
    \item \textbf{GQA}. The GQA API contains all the contents in the API from Listing~\ref{listing} up until the \texttt{llm\_query} function, which is not used. The \texttt{ImagePatch} usage examples look like the following:
\begin{lstlisting}[language=Python, caption=\textbf{GQA example.}, xleftmargin=.0\textwidth, xrightmargin=.0\textwidth]
# Is there a backpack to the right of the man?
def execute_command(image)->str:
    image_patch = ImagePatch(image)
    man_patches = image_patch.find("man")
    # Question assumes one man patch
    if len(man_patches) == 0:
        # If no man is found, query the image directly
        return image_patch.simple_query("Is there a backpack to the right of the man?")
    man_patch = man_patches[0]
    backpack_patches = image_patch.find("backpack")
    # Question assumes one backpack patch
    if len(backpack_patches) == 0:
        return "no"
    for backpack_patch in backpack_patches:
        if backpack_patch.horizontal_center > man_patch.horizontal_center:
            return "yes"
    return "no"
\end{lstlisting}
    \item \textbf{OK-VQA}. The API only uses the \texttt{simple\_query} method from \texttt{ImagePatch}. It additionally uses the \texttt{llm\_query} function. The \texttt{ImagePatch} usage examples look like the following:
\begin{lstlisting}[language=Python, caption=\textbf{OK-VQA example.}, xleftmargin=.0\textwidth, xrightmargin=.0\textwidth]

# Who is famous for allegedly doing this in a lightning storm?
def execute_command(image)->str:
    # The question is not direct perception, so we need to ask the image for more information
    # Salient information: what is being done?
    image = ImagePatch(image)
    guesses = []
    action = image.simple_query("What is being done?")
    external_knowledge_query = "Who is famous for allegedly {} in a lightning storm?".format(action)
    step_by_step_guess = llm_query(external_knowledge_query)
    guesses.append("what is being done is {}".format(action) + ", so " + step_by_step_guess)
    direct_guess = image.simple_query("Who is famous for allegedly doing this in a lightning storm?")
    guesses.append(direct_guess)
    return process_guesses("Who is famous for allegedly doing this in a lightning storm?", guesses)
\end{lstlisting}
\vspace{-0.1cm}
\item \textbf{NeXT-QA}. The \texttt{VideoSegment} class is added to the API definition, and the available \texttt{ImagePatch} methods are \texttt{find}, \texttt{exists}, \texttt{best\_text\_match} and \texttt{simple\_query}. The function \texttt{best\_image\_match} is also used. The \texttt{ImagePatch} usage examples look like:
\begin{lstlisting}[language=Python, caption=\textbf{NeXT-QA example.}, xleftmargin=.0\textwidth, xrightmargin=.0\textwidth]
# why does the man with a red hat put his arm down at the end of the video
# possible answers: ['watching television', 'searching for food', 'move its head', 'looking over cardboard box', 'looks at the camera']
def execute_command(video, possible_answers, question)->[str, dict]:
    # Reason every step
    video_segment = VideoSegment(video)
    # Caption last frame of the video (end of video)
    last_frame = ImagePatch(video_segment, -1)
    last_caption = last_frame.simple_query("What is this?")
    men = last_frame.find("man")
    if len(men) == 0:
        men = [last_frame]
    man = men[0]
    man_action = man.simple_query("What is the man doing?")
    # Answer the question. Remember to create the info dictionary
    info = {
        "Caption of last frame": last_caption,
        "Man looks like he is doing": man_action
    }
    answer = video_segment.select_answer(info, question, possible_answers)
    return answer, info
\end{lstlisting}
    \item \textbf{Beyond benchmarks}. For the examples in Figure~\ref{fig:ood} we use the same API as the one used for the benchmarks, and the usage examples are taken from the benchmark APIs, combining them to have more generality. We do not add any other example, \viper generalizes to the complex cases shown in Figure~\ref{fig:ood} just based on the provided API.
    
\end{itemize}

Note that in some of the examples we added comments, as well as error handling. The generated code also contains similar lines. We removed those for clarity in the figures shown in the main paper.
\end{document}